\pdfoutput=1

\PassOptionsToPackage{table}{xcolor}

\documentclass[11pt]{article}

\usepackage[final]{ACL2023}

\usepackage{times}
\usepackage{latexsym}

\usepackage[T1]{fontenc}

\usepackage[utf8]{inputenc}

\usepackage{microtype}

\usepackage{inconsolata}

\usepackage[table]{xcolor}
\usepackage{multirow}

\usepackage{graphicx}

\usepackage{subcaption}

\usepackage{dblfloatfix}

\usepackage{float}

\newcounter{row}
\makeatletter
\@addtoreset{subfigure}{row}
\makeatother

\title{DrBERT: A Robust Pre-trained Model in French for Biomedical and Clinical domains}

\author{Yanis Labrak{\normalfont $^*$\textsuperscript{1,4}} \hspace{0.5cm} Adrien Bazoge{\normalfont $^*$\textsuperscript{2,3}} \hspace{0.5cm} Richard Dufour{\normalfont \textsuperscript{2}} \hspace{0.5cm} Mickael Rouvier{\normalfont \textsuperscript{1}} \\ {\bf Emmanuel Morin}\textsuperscript{2} \hspace{0.5cm} {\bf Béatrice Daille}\textsuperscript{2} \hspace{0.5cm} {\bf Pierre-Antoine Gourraud}\textsuperscript{3} \\ 
       \textsuperscript{1}LIA, Avignon Université \hspace{0.5cm} \textsuperscript{2}LS2N, UMR CNRS 6004, Nantes Université \\ \textsuperscript{3}Clinique des données, CHU de Nantes, Nantes Université \hspace{0.5cm} \textsuperscript{4}Zenidoc \\
\texttt{\{firstname.lastname\}@univ-avignon.fr} \\ \texttt{\{firstname.lastname\}@univ-nantes.fr} 
}

\begin{document}
\maketitle
\def\thefootnote{*}\footnotetext{Equal contribution.}\def\thefootnote{\arabic{footnote}}
\begin{abstract}
In recent years, pre-trained language models (PLMs) achieve the best performance on a wide range of natural language processing (NLP) tasks. While the first models were trained on general domain data, specialized ones have emerged to more effectively treat specific domains. 
In this paper, we propose an original study of PLMs in the medical domain on French language. We compare, for the first time, the performance of PLMs trained on both public data from the web and private data from healthcare establishments. We also evaluate different learning strategies on a set of biomedical tasks. In particular, we show that we can take advantage of already existing biomedical PLMs in a foreign language by further pre-train it on our targeted data.
Finally, we release the first specialized PLMs for the biomedical field in French, called DrBERT, as well as the largest corpus of medical data under free license on which these models are trained.
\end{abstract}






\section{Introduction}


During the last years, pre-trained language models (PLMs) have been shown to significantly improve performance on many Natural Language Processing (NLP) tasks. 
Recent models, such as BERT~\cite{https://doi.org/10.48550/arxiv.1810.04805} or RoBERTa~\cite{https://doi.org/10.48550/arxiv.1907.11692}, are more and more taking advantage of the huge quantities of unlabeled data thanks to recent  unsupervised approaches like masked language models based on the Transformers architecture~\cite{https://doi.org/10.48550/arxiv.1706.03762}. Most of these PLMs are frequently pre-trained on general domain corpora, such as news articles, books or encyclopedia. An additional step of fine-tuning can also be applied to use these PLMs on a targeted task~\cite{https://doi.org/10.48550/arxiv.1810.04805}. 






Although these generic models are used in various contexts, recent works have shown that optimal performance in specialized domains, such as finance~\cite{https://doi.org/10.48550/arxiv.2006.08097}, medical~\cite{Yang2022} or traveling~\cite{https://doi.org/10.48550/arxiv.2109.01048}, can only be achieved using PLMs adapted to the targeted conditions.


The adaptation of language models to a domain generally follows two strategies. The first is the training {\it from scratch} of a new model using only textual data of the targeted specialty. The second approach, called {\it continual pre-training}~\cite{howard-ruder-2018}, pursues the training of already pre-trained models, allowing them to pass from a generic model to a specialized one. Even if studies have shown that the first strategy generally offers better performance~\cite{Lee_2019}, the second requires a much more constrained number of resources~\cite{https://doi.org/10.48550/arxiv.2010.02559,el-boukkouri-etal-2022} whether in terms of computing resources or of amount of data.



However, domain specific data are generally difficult to obtain, resulting in quite a few specialized PLMs available. This difficulty is even greater for languages other than English.
For the medical domain, data produced during clinical care contain the finesse of medical reasoning and are the most prevalent in terms of quantity. However, they are rarely accessible due to patient privacy constraints.

In this paper, we describe and freely disseminate DrBERT, the first RoBERTa-based PLMs specialized in the biomedical field in French, and the corpus that had allowed their trainings. We also propose an original study concerning the evaluation of different language model pre-training strategies for the medical field, while comparing it with our model derived from clinical private data, called ChuBERT. In our experiments, PLMs with publicly available biomedical data can result in similar or better performance compared to highly specialized private data collected from hospital reports or to larger corpora having only generic data. Our contributions can be summarized as follows:

\begin{itemize}

\item A new benchmark aggregating a set of NLP tasks in the medical field in French has been set up, making it possible to evaluate language models at the syntactic and semantic level (multi-label classification, part-of-speech tagging, named entity recognition, etc.). 

\item A large textual data collection, called NACHOS, crawled from multiple biomedical online sources.

\item The construction and evaluation of the first open-source PLMs in French for the biomedical domain based on RoBERTa architecture, called DrBERT, including the analysis of different pre-training strategies.

\item A set of models using both public and private data trained on comparable data sizes. These models were then compared by evaluating their performance on a wide range of tasks, both public and private.

\item The free distribution of the public PLMs under the open-source MIT license as well as the NACHOS corpus under CC0 1.0 license\footnote{\href{https://drbert.univ-avignon.fr/}{https://drbert.univ-avignon.fr/}}.





\end{itemize}

\section{Related work}






BERT~\cite{https://doi.org/10.48550/arxiv.1810.04805} is a contextualized word representation model based on the concept of masked language model and pre-trained using bidirectional Transformers~\cite{https://doi.org/10.48550/arxiv.1706.03762}. 
Since its release, it obtains state-of-the-art (SOTA) performance on almost every NLP tasks, while requiring minimal task-specific architectural modifications.


However, the training cost of such a model is very high in terms of computation due to the complexity of each training objective and the quantity of data needed. Consequently, new methods emerge and propose more effective ways of performing pre-training. One of them is RoBERTa~\cite{https://doi.org/10.48550/arxiv.1907.11692}. In order to improve the initial BERT model, the authors made some simple design changes in its training procedure. They modify the masked-language strategy to perform dynamic masking, remove the next sentence prediction task, increase dramatically the batch sizes and use significantly more data during a longer training period. Nowadays, RoBERTa is the standard model for a lot of NLP tasks and languages, including French with CamemBERT model~\cite{Martin_2020}.


Recently, multiple language models have been developed for biomedical and clinical fields through unsupervised pre-training of Transformer-based architectures, mainly for English language. One of the first models was BioBERT~\cite{Lee_2019}, which is based on the initially pre-trained BERT model and further pre-trained using biomedical-specific data through continual pre-training. Other models like BlueBERT~\cite{peng2019transfer} and ClinicalBERT~\cite{https://doi.org/10.48550/arxiv.1904.05342} also used this approach on various data sources. An alternative method, when enough in-domain data is available, is to directly pre-train models from scratch (SciBERT~\cite{Beltagy2019SciBERT}, PubMedBERT~\cite{Gu_2021}, etc.). Note that SciBERT was trained on mixed-domain data from biomedical and computer science domains, while PubMedBERT on biomedical data only. ~\citet{Gu_2021} disputed the benefits of mixed-domain data for pre-training, based on results obtained on tasks from BLURB benchmark.



In other languages than English, BERT-based models are much rarer and primarily rely on continual pre-training. Examples include German~\cite{Shrestha2021}, Portuguese~\cite{schneider-etal-2020-biobertpt}, and Swedish~\cite{vakili-etal-2022-downstream}. Only the Spanish~\cite{https://doi.org/10.48550/arxiv.2109.03570} and Turkish~\cite{Hazal_2022} models were trained from scratch with biomedical and clinical data from various sources. For French, there is, to our knowledge, no publicly available model specifically built for the biomedical domain.

















\section{Pre-training datasets}
\label{s:datasets}

In the biomedical domain, previous works~\cite{Gu_2021} on PLMs highlighted the importance of matching the data sources used for its training to the targeted downstream tasks. 
Due to their sensitive nature (protection of user data, protected health information of patients, etc.), medical data are extremely difficult to obtain. Massive collection of web data related to this domain appears to be a solution that can overcome this lack. However, these web documents vary in terms of quality. No comparison has been made between PLMs based on specific domain data from the web and those on private documents from clinical data warehouses, whose quality can be controlled.

We extracted two different medical datasets for French. The first one gathers data crawled from a variety of free-of-use online sources, and the second one private hospital stays reports from the Nantes University Hospital. 

Table~\ref{t:resume_corpus} gives a general overview of the two collected corpora. The public web-based data, detailed in Section~\ref{public-data-nachos}, allowed the constitution of a corpus, called NACHOS$_{large}$, containing 7.4 GB of data. The private dataset, called NBDW$_{small}$ is described in Section~\ref{private-data} and contains 4 GB of data. In order to perform comparable experiments, we extracted a NACHOS sub-corpus (NACHOS$_{small}$) of the same size as the private data. Finally, Section~\ref{s:preprocess} describes the pre-processing applied to both datasets.

\begin{table}[H]
\footnotesize
\centering
\begin{tabular}{llll}
\hline
Corpus & \multicolumn{1}{l}{Size} & \multicolumn{1}{l}{\#words} & \multicolumn{1}{l}{\#sentences} \\ \hline
NACHOS$_{large}$ (pub.) & 7.4 GB & 1.1 B & 54.2 M \\
NACHOS$_{small}$ (pub.) & 4 GB & 646 M & 25.3 M \\\hline
NBDW$_{small}$ (private) & 4 GB & 655 M & 43.1 M \\
NBDW$_{mixed}$ (both) & 4+4 GB & 1.3 B & 68.4 M \\ \hline
\end{tabular}
\caption{\label{t:resume_corpus}Overview of the public (NACHOS) and private (NBDW) collected datasets.}
\end{table}

\subsection{Public corpus - NACHOS}
\label{public-data-nachos}

We introduce the opeN crAwled frenCh Healthcare cOrpuS (NACHOS), a French medical open-source dataset compiled by crawling a variety of textual sources around the medical topic. It consists of more than one billion words, drawn from 24 French speaking high-quality websites. The corpus includes a wide range of medical information: descriptions of diseases and conditions, information on treatments and medications, general health-related advice, official scientific meeting reports, anonymized clinical cases, scientific literature, thesis, French translation pairs, university health courses and a large range of data obtained from raw textual sources, web scrapping, and optical character recognition (OCR). Table~\ref{table:nachos-sources} summarizes the different data sources of NACHOS.

We use heuristics to split the texts into sentences and aggressively filter out short or low-quality sentences like those obtained from OCR. Finally, we classified them into languages by using our own classifier trained on the multilingual Opus EMEA~\cite{tiedemann-nygaard-2004-opus} and MASSIVE~\cite{https://doi.org/10.48550/arxiv.2204.08582} corpora to keep only the sentences in French.



For the 4 GB version of NACHOS (NACHOS$_{small}$), we shuffled the whole corpus and selected randomly 25.3M sentences to maximize data sources homogeneity.
The full NACHOS corpus is now freely available online\footnote{\href{https://drbert.univ-avignon.fr/}{https://drbert.univ-avignon.fr/}}.

\begin{table}[H]
\footnotesize
\centering
\center
\begin{tabular}{lr}
\hline
\textbf{Resource name}           & \textbf{\# words}  \\ \hline
HAL                              & 638,508,261            \\
Haute Autorité de Santé (HAS)    & 113,394,539            \\
Drug leaflets                    & 74,770,229             \\
Medical Websites Scrapping       & 60,561,495             \\
ANSES SAISINE                    & 51,372,932             \\
Public Drug Database (BDPM)      & 48,302,695             \\
ISTEX                            & 44,124,422             \\
CRTT                             & 26,210,756             \\
WMT-16                           & 10,282,494             \\
EMEA-V3                          & 6,601,617              \\
Wikipedia Life Science French    & 4,671,944              \\
Cochrane summaries               & 3,932,526              \\
ANSES RCP                        & 2,953,045              \\
Cerimes                          & 1,717,552              \\
Vidal                            & 410,313                \\
LiSSa                            & 235,838                \\
DEFT-2020                        & 231,396                \\
CLEAR                            & 225,898                \\
CNEDiMTS                         & 175,416                \\
QUAERO French Medical Corpus     & 72,031                 \\
ANSM Clinical Study Registry     & 47,678                 \\
ECDC                             & 44,482                 \\
QualiScope                       & 12,718                 \\
WMT-18-Medline                   & 7,673                  \\ \hline
\textbf{Total}                   & \textbf{1,088,867,950} \\ \hline
\end{tabular}
\caption{Sources of the NACHOS corpus.}
\label{table:nachos-sources}
\end{table}


\subsection{Private corpus - NBDW}
\label{private-data}

The private corpus, called Nantes Biomedical Data Warehouse (NBDW), was obtained using the data warehouse from Nantes University Hospital. This data warehouse includes different dimensions of patients' related data: socio-demographic, drug prescriptions and other information associated with consultation or hospital stays (diagnosis, biology, imagery, etc.). The authorization to implement and exploit the NBDW dataset was granted in 2018 by the CNIL ({\it Commission National de l'Informatique et des Libertés}), the French independent supervisory authority in charge of application of national and European data privacy protection laws; authorization N°2129203.

For this work, a sample of 1.7 million de-identified hospital stays reports was randomly selected and extracted from the data warehouse. As described in Table~\ref{table:Medical-Spec-ChuBERT}, the reports are from various hospital departments, emergency medicine, gynecology and ambulatory care being the most frequent.


Each reports was split into tokens sequence with an average of 15.26 words per sequence. Then, all tokens sequences from all reports were shuffled to build the corpus.
This corpus contains 655M words, from 43.1M sentences, for a total size of approximately 4 GB. 

\begin{table}[H]
\footnotesize
\centering
\center
\begin{tabular}{lrr}
\hline
\textbf{Medical Specialty}                 & \textbf{\# documents}       & \textbf{\# words} \\ \hline
Other                                      & 474,588                         & 192,832,792           \\
Emergency Medicine                         & 235,579                         & 90,807,406            \\
Ambulatory Care                            & 119,149                         & 50,975,472            \\
Consultation                               & 95,135                          & 38,335,804            \\
Gynecology                                 & 132,983                         & 38,204,495            \\
Cardiology                                 & 29,633                          & 22,654,583            \\
Medical Oncology                           & 45,603                          & 22,587,869            \\
Gastroenterology                           & 46,600                          & 21,340,794            \\
Orthopaedic Surgery                        & 82,084                          & 18,983,791            \\
Hematology                                 & 41,776                          & 18,285,983            \\
Critical Care Medicine                     & 20,819                          & 16,472,785            \\
Otolaryngology                             & 69,343                          & 16,131,214            \\
Dermatology                                & 51,804                          & 15,035,412            \\
Rheumatology                               & 31,527                          & 14,647,543            \\
Urology                                    & 51,535                          & 14,272,231            \\
Colon and Rectal Surgery                   & 45,987                          & 13,334,550            \\
Internal Medicine                          & 23,904                          & 13,282,253            \\
Psychiatry                                 & 26,628                          & 12,496,503            \\
Neurosurgery                               & 34,481                          & 10,360,533            \\
Nephrology                                 & 19,171                          & 9,548,533             \\
Ophthalmology                               & 19,700                          & 4,464,515             \\ \hline
\textbf{Total}                             & \textbf{1,698,029}              & \textbf{655,055,061}  \\ \hline
\end{tabular}
\caption{Sources of the NBDW corpus.}
\label{table:Medical-Spec-ChuBERT}
\end{table}

\subsection{Pre-processing step}
\label{s:preprocess}

The supplied text data has been split into subword units using  SentencePiece~\cite{https://doi.org/10.48550/arxiv.1808.06226}, an extension of Byte-Pair encoding (BPE)~\cite{sennrich-etal-2016-neural} and WordPiece~\cite{https://doi.org/10.48550/arxiv.1609.08144} that does not require pre-tokenization (at the word or token level), thereby avoiding the requirement for language-specific tokenizers. We employ a vocabulary size of 32k subword tokens. For each model pre-trained from scratch (see Section~\ref{s:strategies}), tokenizers were built using all the sentences from the pre-training dataset.


\section{Models pre-training}
\label{models:pre-training-strategies}



In this section, we describe the pre-training modalities of our studied models from two points of view: 1) the influence of the data used (size and nature), and 2) the pre-training strategies of the models. These two levels are respectively detailed in Sections~\ref{s:data} and~\ref{s:strategies}. Section~\ref{s:sota_models} finally presents the existing state-of-the-art pre-trained models that will be used for comparison purposes.

\subsection{Influence of data}
\label{s:data}

One issue is to identify the amount of data required to create a model that performs well and can compete with models trained on general domains. Recent studies, such as those by \citet{https://doi.org/10.48550/arxiv.2011.04946} and \citet{Martin_2020}, discuss the impact of the size of pre-training data on model performance. 
According to these studies, some tasks are performing better with fewer data while others, such as commonsense knowledge and reasoning tasks, keep improving performance when pre-training data are added. 




In the medical field, no study has been conducted to compare the impact of varying the amount of domain-specific data during pre-training, or to assess the impact of the supposedly variable quality of the data depending on their source of collection.

We thus propose to evaluate the pre-training of several language models on either NACHOS$_{small}$ or NBDW$_{small}$ corpus, as described in Section~\ref{s:datasets}. Additionally, we propose a model pre-trained on NACHOS$_{large}$ to investigate if having almost twice as much data improves model performance. Finally, a combination of both public NACHOS$_{small}$ and NBDW$_{small}$ sources for a total of 8 GB (NBDW$_{mixed}$) is explored, to demonstrate if combining private and public data is a viable approach in low-resource domains.

\subsection{Pre-training strategies}
\label{s:strategies}

In addition to the analysis on the size and the sources of data, we also seek to evaluate three training strategies of PLMs for the medical domain:
\begin{itemize}

  \item Training a full model from scratch, including the subword tokenizer.
  
  \item Continuing the pre-training of the state-of-the-art language model for French, called CamemBERT, on our medical-specific data while keeping the initial tokenizer.
  
  \item Continuing the pre-training of a state-of-the-art domain specific language model for medical but here in English, called PubMedBERT, on our French data while keeping the initial tokenizer.
  
\end{itemize}

Regarding the last strategy, our objective is to compare the performance of an English medical model further pre-trained on our French medical data, against another one based on a generic French model. Indeed, the medical domains shares many terms across languages that make relevant the mixture of resources from two languages.

Table~\ref{table:pretrained-models} summarizes all the configurations evaluated in this paper, integrating both the study of data size and pre-training strategies.

\begin{table}[H]
\scriptsize
\setlength\extrarowheight{1pt}
\centering
\center
\begin{tabular}{lcl}
\hline
Model name & Strategy & Corpus \\ \hline
DrBERT & From scratch & NACHOS$_{large}$ \\
DrBERT & From scratch & NACHOS$_{small}$ \\ 
ChuBERT & From scratch & NBDW$_{small}$ \\
ChuBERT & From scratch & NBDW$_{mixed}$  \\ \hline

CamemBERT & continual pre-training & NACHOS$_{small}$ \\
PubMedBERT & continual pre-training & NACHOS$_{small}$ \\ 
CamemBERT & continual pre-training & NBDW$_{small}$\\ \hline
\end{tabular}
\caption{List of studied pre-trained model configurations.}
\label{table:pretrained-models}
\end{table}

\paragraph{Model architecture} All models pre-trained from-scratch use the CamemBERT~\textsubscript{base} configuration, which is the same as RoBERTa~\textsubscript{base} architecture (12 layers, 768 hidden dimensions, 12 attention heads, 110M parameters). We did not train the large version of our models due to resources limitations.

\paragraph{Language modeling} We train the models on the Masked Language Modeling (MLM) task using HuggingFace library~\cite{https://doi.org/10.48550/arxiv.1910.03771}. It consists of randomly replacing a subset of tokens from the sequence by a special token, and asking the model to predict them using cross-entropy loss. In BERT and RoBERTa models (including CamemBERT), 15\% of the tokens are randomly selected. Of those selected tokens, 80\% are replaced with the \texttt{<mask>} token, 10\% remain unchanged and 10\% are randomly replaced by a token from the vocabulary. We keep this masking probability of 15\% for the training of our models.

\paragraph{Optimization \& Pre-training} We optimize the models for 80k steps with batch sizes of 4,096 sequences, each sequence filled with 512 tokens, allowing to process 2.1M tokens per step. The learning rate is warmed up linearly for 10k steps, going up from zero to the initial 5$\times$10$\textsuperscript{-5}$ learning rate. Models are trained on 128 Nvidia V100 32 GB GPUs for 20 hours on Jean Zay supercomputer.
We use mixed precision training (FP16)~\cite{https://doi.org/10.48550/arxiv.1710.03740} to reduce the memory footprint, allowing us to enlarge the batch size to 32 sequences on each GPU.

\subsection{Baseline models}
\label{s:sota_models}

We describe some existing pre-trained models used as baselines in our comparative study.

{\bf CamemBERT}~\cite{Martin_2020} is a RoBERTa based model pre-trained totally from scratch on the French subset of OSCAR corpus (138 GB). In our case, this model is our main baseline to compare our results on, since it is the state-of-the-art model for French. We also use the 4 GB model's variants of CamemBERT to compare the impact of the nature and quantity of the data.

{\bf PubMedBERT}~\cite{Gu_2021} is a BERT based biomedical-specific model pre-trained totally from scratch on the 3.1 billions words of PubMed corpus (21 GB).


{\bf ClinicalBERT}~\cite{https://doi.org/10.48550/arxiv.1904.05342} is a clinical-specific model based on BERT tokenizer and weights, which has been further pre-trained on the 0.5 billion words of MIMIC corpus (3.7 GB).

{\bf BioBERT v1.1}~\cite{Lee_2019} is a biomedical-specific model based on BERT tokenizer and weights which has been further pre-trained using the 4.5 billion words of PubMed corpus.


\section{Downstream evaluation tasks}

\begin{table*}[ht]
\scriptsize
\centering
\center
\begin{tabular}{cccccc}
\hline
\textbf{Thematic / Corpus name} & \textbf{Task} & \textbf{Metric} & \textbf{Train} & \textbf{Dev} & \textbf{Test} \\ \hline

\multicolumn{3}{c}{\textit{Public Corpus}} \\
\rule{0pt}{3ex}

ESSAIS \cite{dalloux_claveau_grabar_oliveira_moro_gumiel_carvalho_2021} & POS Tagging & Macro F1 & 9,693 & 2,077 & 2,078 \\
CAS: French Corpus with Clinical Cases \cite{grabar:hal-01937096} & POS Tagging & Macro F1 & 5,306 & 1,137  & 1,137 \\
MUSCA-DET - Social Determinants of Health extraction (Task 1) & Nested NER & Macro F1 & 19,861 & 2,207 & 5,518 \\
MUSCA-DET - Social Determinants of Health extraction (Task 2) & Multi-label Classification & Macro F1 & 19,861 & 2,207 & 5,518 \\
QUAERO French Medical Corpus - EMEA~\cite{Nvol2014TheQF} & Nested NER & Weighted F1 & 11 & 12 & 15 \\
QUAERO French Medical Corpus - MEDLINE~\cite{Nvol2014TheQF} & Nested NER & Weighted F1 & 833 & 832 & 833 \\
FrenchMedMCQA~\cite{labrak:hal-03824241} & MCQA & EMR / Hamming Score & 2,171 & 312 & 622 \\ \hline

\multicolumn{3}{c}{\textit{Private Corpus}} \\
\rule{0pt}{3ex}

Medical report acute heart failure structuration & Named Entity Recognition & Macro F1 & 2,527 & 281 & 703  \\
Acute heart failure (aHF) classification & Binary Classification & Macro F1 & 1,179 & 132 & 328  \\

Technical Specialties Sorting & Classification Multi-class & Macro F1 & 4,413 & 1,470 & 1,473  \\
Medical report structuration prescriptions & Named Entity Recognition & Macro F1 & 61 & 15 & 26 \\ \hline


\end{tabular}%
\caption{Corpus, tasks and metrics synthesis for evaluating medical-specific models.\label{fig:synthese}}
\end{table*}

To evaluate the different pre-training configurations of our models, a set of tasks in the medical domain is necessary. While this NLP domain-specific benchmark exists in English (BLURB~\cite{Gu_2021}), none exist for French. In this section, we describe an original benchmark, summarized in Table~\ref{fig:synthese}, integrating various NLP medical tasks for French. Among them, some are from publicly-available datasets (Section~\ref{s:public_tasks}), allowing the replication of our experiments. Other tasks come from private datasets (Section~\ref{s:private_tasks}) and cannot be shared. However, they are useful to evaluate our models more accurately.

\subsection{Publicly-available tasks}
\label{s:public_tasks}

\paragraph{ESSAIS / CAS: French Corpus with Clinical Cases} The ESSAIS~\cite{dalloux_claveau_grabar_oliveira_moro_gumiel_carvalho_2021} and CAS~\cite{grabar:hal-01937096} corpora respectively contain 13,848 and 7,580 clinical cases in French. Some clinical cases are associated with discussions. A subset of the whole set of cases is enriched with morpho-syntactic (part-of-speech (POS) tagging, lemmatization) and semantic (UMLS concepts, negation, uncertainty) annotations. In our case, we focus only on the POS tagging task.

\paragraph{FrenchMedMCQA} The FrenchMedMCQA corpus~\cite{labrak:hal-03824241} is a publicly available Multiple-Choice Question Answering (MCQA) dataset in French for medical domain. It contains 3,105 questions coming from real exams of the French medical specialization diploma in pharmacy, integrating single and multiple answers.

\paragraph{QUAERO French Medical Corpus} The QUAERO French Medical Corpus~\cite{Nvol2014TheQF} introduces an extensive corpus of biomedical documents annotated at the entity and concept levels to provide NER and classification tasks. Three text genres are covered, comprising a total of 103,056 words obtained either from EMEA or MEDLINE. Ten entity categories corresponding to UMLS~\cite{UMLS_Bodenreider} Semantic Groups were annotated, using automatic pre-annotations validated by trained human annotators. Overall, a total of 26,409 entity annotations were mapped to 5,797 unique UMLS concepts. To simplify the evaluation process, we sort the nested labels by alphabetical order and concatenate them together into a single one to transform the task to a usable format for token classification with BERT based architectures.

\paragraph{MUSCA-DET} MUSCA-DET is a French corpus of sentences extracted from the "Lifestyle" section in clinical notes from Nantes University Hospital biomedical data warehouse. The corpus contains 27,000 pseudonymized sentences annotated with 26 entities related to Social Determinants of Health (living, marital status, housing, descendants, employment, alcohol, smoking, drug abuse, physical activity). The corpus includes two tasks: nested name entity recognition (NER) and multi-label classification.

\begin{table*}[!htb]
\tiny
\setlength\extrarowheight{1pt}
\centering
\center
\begin{tabular}{l ccc ccc ccc ccc}
\hline
 &
  \multicolumn{3}{c}{\textbf{aHF NER}} & 
  \multicolumn{3}{c}{\textbf{aHF classification}} & 
  \multicolumn{3}{c}{\textbf{NER Medical Report}} &
  \multicolumn{3}{c}{\textbf{Specialities Classification}} \\ \hline
 &
  \multicolumn{1}{c}{\textit{\textbf{P}}} &
  \multicolumn{1}{c}{\textit{\textbf{R}}} &
  \textit{\textbf{F1}} &
  \multicolumn{1}{c}{\textit{\textbf{P}}} &
  \multicolumn{1}{c}{\textit{\textbf{R}}} &
  \textit{\textbf{F1}} &
  \multicolumn{1}{c}{\textit{\textbf{P}}} &
  \multicolumn{1}{c}{\textit{\textbf{R}}} &
  \textit{\textbf{F1}} &
  \multicolumn{1}{c}{\textit{\textbf{P}}} &
  \multicolumn{1}{c}{\textit{\textbf{R}}} &
  \textit{\textbf{F1}} \\ \hline
\textbf{CamemBERT OSCAR 138 GB} &
  \multicolumn{1}{c}{40.89} &
  \multicolumn{1}{c}{35.22} &
  35.13 &
  \multicolumn{1}{c}{81.90} &
  \multicolumn{1}{c}{79.12} &
  80.13 &
  \multicolumn{1}{c}{87.98} &
  \multicolumn{1}{c}{91.66} &
  89.35 &
  \multicolumn{1}{c}{99.32} &
  \multicolumn{1}{c}{99.09} &
  99.20 \\
\textbf{CamemBERT OSCAR 4 GB} &
  \multicolumn{1}{c}{46.32} &
  \multicolumn{1}{c}{43.17} &
  42.66 &
  \multicolumn{1}{c}{81.49} &
  \multicolumn{1}{c}{81.42} &
  81.41 &
  \multicolumn{1}{c}{87.79} &
  \multicolumn{1}{c}{90.74} &
  88.78 &
  \multicolumn{1}{c}{99.53} &
  \multicolumn{1}{c}{99.69} &
  99.61 \\
\textbf{CamemBERT CCNET 4 GB} &
  \multicolumn{1}{c}{47.25} &
  \multicolumn{1}{c}{42.2} &
  43.11 &
  \multicolumn{1}{c}{82.02} &
  \multicolumn{1}{c}{79.30} &
  79.98 &
  \multicolumn{1}{c}{87.61} &
  \multicolumn{1}{c}{92.28} &
  89.34 &
  \multicolumn{1}{c}{99.54} &
  \multicolumn{1}{c}{99.55} &
  99.55 \\ 
\textbf{PubMedBERT} &
  \multicolumn{1}{c}{52.61} &
  \multicolumn{1}{c}{46.30} &
  47.22 &
  \multicolumn{1}{c}{78.17} &
  \multicolumn{1}{c}{76.18} &
  76.86 &
  \multicolumn{1}{c}{87.07} &
  \multicolumn{1}{c}{92.61} &
  89.20 &
  \multicolumn{1}{c}{99.25} &
  \multicolumn{1}{c}{99.51} &
  99.37 \\
\textbf{ClinicalBERT} &
  \multicolumn{1}{c}{50.11} &
  \multicolumn{1}{c}{44.15} &
  44.70 &
  \multicolumn{1}{c}{80.13} &
  \multicolumn{1}{c}{75.92} &
  77.12 &
  \multicolumn{1}{c}{87.04} &
  \multicolumn{1}{c}{92.14} &
  88.77 &
  \multicolumn{1}{c}{98.58} &
  \multicolumn{1}{c}{98.62} &
  98.58 \\
\textbf{BioBERT v1.1} &
  \multicolumn{1}{c}{49.37} &
  \multicolumn{1}{c}{47.25} &
  46.01 &
  \multicolumn{1}{c}{79.69} &
  \multicolumn{1}{c}{78.51} &
  79.00 &
  \multicolumn{1}{c}{\textbf{88.17}} &
  \multicolumn{1}{c}{91.80} &
  89.38 &
  \multicolumn{1}{c}{98.59} &
  \multicolumn{1}{c}{99.03} &
  98.80 \\ \hline
\textbf{DrBERT NACHOS$_{large}$} &
  \multicolumn{1}{c}{\underline{55.29}} &
  \multicolumn{1}{c}{46.66} &
  48.22 &
  \multicolumn{1}{c}{81.33} &
  \multicolumn{1}{c}{81.25} &
  81.25 &
  \multicolumn{1}{c}{\underline{87.99}} &
  \multicolumn{1}{c}{\textbf{92.80}} &
  \textbf{89.83} &
  \multicolumn{1}{c}{\underline{99.82}} &
  \multicolumn{1}{c}{\textbf{99.90}} &
  \textbf{99.86} \\
\textbf{DrBERT NACHOS$_{small}$} &
  \multicolumn{1}{c}{54.55} &
  \multicolumn{1}{c}{43.39} &
  45.93 &
  \multicolumn{1}{c}{79.85} &
  \multicolumn{1}{c}{80.10} &
  79.87 &
  \multicolumn{1}{c}{87.57} &
  \multicolumn{1}{c}{\underline{92.76}} &
  89.44 &
  \multicolumn{1}{c}{\textbf{99.85}} &
  \multicolumn{1}{c}{\underline{99.85}} &
  \underline{99.85} \\
\textbf{ChuBERT NBDW$_{small}$} &
  \multicolumn{1}{c}{\textbf{56.92}} &
  \multicolumn{1}{c}{47.46} &
  \underline{49.01} &
  \multicolumn{1}{c}{81.03} &
  \multicolumn{1}{c}{\textbf{82.67}} &
  \underline{81.56} &
  \multicolumn{1}{c}{87.76} &
  \multicolumn{1}{c}{92.63} &
  \underline{89.58} &
  \multicolumn{1}{c}{99.76} &
  \multicolumn{1}{c}{\textbf{99.90}} &
  99.83 \\
\textbf{ChuBERT NBDW$_{mixed}$} &
  \multicolumn{1}{c}{54.62} &
  \multicolumn{1}{c}{\underline{47.81}} &
  \textbf{49.14} &
  \multicolumn{1}{c}{\underline{82.23}} &
  \multicolumn{1}{c}{\underline{81.71}} &
  \textbf{81.98} &
  \multicolumn{1}{c}{87.42} &
  \multicolumn{1}{c}{92.36} &
  89.30 &
  \multicolumn{1}{c}{99.81} &
  \multicolumn{1}{c}{99.82} &
  99.81 \\ \hline
\textbf{CamemBERT NACHOS$_{small}$} &
  \multicolumn{1}{c}{22.02} &
  \multicolumn{1}{c}{16.67} &
  16.08 &
  \multicolumn{1}{c}{74.86} &
  \multicolumn{1}{c}{69.82} &
  69.80 &
  \multicolumn{1}{c}{65.72} &
  \multicolumn{1}{c}{68.49} &
  66.74 &
  \multicolumn{1}{c}{99.44} &
  \multicolumn{1}{c}{99.67} &
  99.54 \\
\textbf{PubMedBERT NACHOS$_{small}$} &
  \multicolumn{1}{c}{53.44} &
  \multicolumn{1}{c}{\textbf{48.21}} &
  48.72 &
  \multicolumn{1}{c}{\textbf{83.06}} &
  \multicolumn{1}{c}{80.39} &
  81.40 &
  \multicolumn{1}{c}{87.35} &
  \multicolumn{1}{c}{92.69} &
  89.36 &
  \multicolumn{1}{c}{99.52} &
  \multicolumn{1}{c}{99.58} &
  99.55 \\
\textbf{CamemBERT NBDW$_{small}$} &
  \multicolumn{1}{c}{25.44} &
  \multicolumn{1}{c}{19.33} &
  19.12 &
  \multicolumn{1}{c}{79.50} &
  \multicolumn{1}{c}{74.74} &
  76.02 &
  \multicolumn{1}{c}{68.80} &
  \multicolumn{1}{c}{71.23} &
  69.64 &
  \multicolumn{1}{c}{99.60} &
  \multicolumn{1}{c}{99.57} &
  99.58 \\ \hline
\end{tabular}
\caption{Performance on our private biomedical downstream tasks. Best model in bold and second is  underlined.\label{fig:downstreamtaskprivate}}
\end{table*}

\subsection{Private tasks}
\label{s:private_tasks}

\paragraph{Technical Specialties Sorting} This classification task has to assign the specialty of a medical of a medical report based on its transcription. The dataset consists of 7,356 French medical reports that have been manually annotated and equally sampled across 6 specialties: Psychiatry, Urology, Endocrinology, Cardiology, Diabetology, and Infectiology.


\paragraph{Medical report structuration prescriptions (NER)} The task seeks to identify named entities in a gold sample of 100 long medical reports obtained from French speech transcriptions. The named entities are annotated using the BIO format and fall into 12 classes: {\it O}, {\it AGE}, {\it CITY}, {\it DATE}, {\it EMAIL}, {\it HOSPITAL}, {\it PHONE}, {\it DOSAGE}, {\it DURATION}, {\it FORM}, {\it MEDICATION} and {\it POSOLOGY}.

\paragraph{Medical report acute heart failure structuration (NER)} This corpus contains 350 hospital stay reports (divided into 3,511 sentences) from Nantes University Hospital. The reports are annotated with 46 entity types related to the following clinical information: cause of chronic heart failure, triggering factor for acute heart failure, diabetes, smoking status, heart rate, blood pressure, weight, height, medical treatment, hypertension and left ventricular ejection fraction. Overall, the corpus contains 6,116 clinical entities.

\paragraph{Acute heart failure (aHF) classification} This task consists of the classification of hospital stays reports according to the presence or absence of a diagnostic of acute heart failure. This corpus consists of 1,639 hospital stays reports from Nantes university hospital, which are labeled as positive or negative to acute heart failure. 













\section{Results and Discussions}


As previously described, we evaluate the performance of our pre-trained language models proposed for the biomedical domain on a set of public and private NLP downstream tasks related to the medical domain. We first propose to analyze the results according to the different pre-training strategies used (Section~\ref{s:pre-train-impact}) then to focus on the impact of the pre-training data, whether in terms of size or nature (Section~\ref{s:data-effect}). Finally, we are interested in the generalization capacities of our domain-specific models by applying and comparing them on general domain NLP tasks (Section~\ref{s:general-domain}).

Note that all the PLMs have been fine-tuned in the same way for all downstream tasks and all the reported results are obtained by averaging the scores from four runs. Performance on biomedical downstream tasks are reported in Tables~\ref{fig:downstreamtaskprivate} and~\ref{fig:downstreamtaskpublic} for respectively private and public tasks. For readability reasons, the first part of each table presents the existing baseline models results, the second part our specialized models trained from-scratch, and the last part our models using continual pre-training.

\begin{table*}[!htb]
\setlength\tabcolsep{1.7pt}
\setlength\extrarowheight{2pt}
\tiny
\centering
\center
\begin{tabular}{l ccc ccc ccc ccc cc ccc ccc}
\hline
 &
  \multicolumn{3}{c}{\textbf{MUSCA-DET T1}} &
  \multicolumn{3}{c}{\textbf{MUSCA-DET T2}} &
  \multicolumn{3}{c}{\textbf{ESSAI POS}} &
  \multicolumn{3}{c}{\textbf{CAS POS}} &
  \multicolumn{2}{c}{\textbf{FrenchMedMCQA}} &
  \multicolumn{3}{c}{\textbf{QUAERO-EMEA}} &
  \multicolumn{3}{c}{\textbf{QUAERO-MEDLINE}}\\ \hline
 &
  \textit{\textbf{P}} &
  \textit{\textbf{R}} &
  \textit{\textbf{F1}} &
  \textit{\textbf{P}} &
  \textit{\textbf{R}} &
  \textit{\textbf{F1}} &
  \textit{\textbf{P}} &
  \textit{\textbf{R}} &
  \textit{\textbf{F1}} &
  \textit{\textbf{P}} &
  \textit{\textbf{R}} &
  \textit{\textbf{F1}} &
  \textit{\textbf{Hamming}} &
  \textit{\textbf{EMR}} &
  \textit{\textbf{P}} &
  \textit{\textbf{R}} &
  \textit{\textbf{F1}} &
  \textit{\textbf{P}} &
  \textit{\textbf{R}} &
  \textit{\textbf{F1}} \\ \hline 
\textbf{CamemBERT OSCAR 138 GB} &
  89.04 & 88.59 &
  88.54 & 89.87 &
  87.12 & 88.20 &
  81.57 & 81.01 &
  81.10 & 96.37 &
  94.53 & 95.22 &
  36.24 & \textbf{16.55} &
  90.57 & 91.06 &
  90.71 & 76.58 &
  78.67 & 77.41 \\
\textbf{CamemBERT OSCAR 4 GB} &
  86.09 & 85.45 &
  85.43 & 92.68 &
  90.34 & 91.27 &
  84.01 & 83.51 &
  83.69 & \underline{98.15} &
  95.34 & 96.42 &
  35.75 & 15.37 &
  90.75 & 91.16 &
  90.83 & 78.55 &
  79.33 & \underline{78.76} \\
\textbf{CamemBERT CCNET 4 GB} &
  91.12 & 89.91 &
  90.33 & 93.10 &
  \underline{90.42} & 91.38 &
  85.60 & 85.63 &
  85.42 & \textbf{98.19} &
  \textbf{96.75} & \textbf{97.33} &
  34.71 & 14.41 &
  90.31 & 90.59 &
  90.33 & 78.06 &
  78.11 & 77.61 \\ 
\textbf{PubMedBERT} &
  93.04 & 91.45 &
  91.99 & 84.41 &
  80.60 & 81.97 &
  88.43 & 87.93 &
  87.78 & 97.40 &
  94.86 & 95.90 &
  33.98 & 14.14 &
  86.89 & 87.33 &
  86.79 & 77.33 &
  77.28 & 77.09 \\
\textbf{ClinicalBERT} &
  91.79 & 89.44 &
  90.36 & 85.43 &
  81.23 & 82.95 &
  89.09 & \underline{88.78} &
  88.24 & 97.94 &
  95.88 & 96.73 &
  32.78 & 14.19 &
  84.91 & 85.47 &
  84.79 & 75.56 &
  74.85 & 75.05 \\
\textbf{BioBERT 1.1} &
  91.82 & 89.82 &
  90.46 & 85.52 &
  80.14 & 81.91 &
  86.76 & 84.90 &
  85.18 & 98.10 &
  \underline{96.39} & \underline{97.12} &
  36.19 & \underline{15.43} &
  84.55 & 85.03 &
  84.29 & 72.62 &
  73.30 & 72.68 \\ \hline
\textbf{DrBERT NACHOS$_{large}$} &
  92.10 & 90.27 &
  91.04 & \textbf{94.97} &
  90.41 & \underline{92.24} &
  \textbf{90.96} & \textbf{89.19} &
  \textbf{89.75} & 97.37 &
  94.49 & 95.65 &
  \underline{36.66} & 15.32 &
  \textbf{91.93} & \textbf{92.52} &
  \textbf{92.09} & 77.85 &
  78.54 & 77.88 \\
\textbf{DrBERT NACHOS$_{small}$} &
  93.35 & 90.62 &
  91.77 & 91.31 &
  86.60 & 88.57 &
  \underline{90.12} & 88.37 &
  \underline{88.76} & 97.04 &
  94.88 & 95.70 &
  \textbf{37.37} & 13.34 &
  \underline{91.54} & \underline{92.00} &
  \underline{91.66} & 77.91 &
  \underline{79.34} & 78.18 \\
\textbf{ChuBERT NBDW$_{small}$} &
  \textbf{94.88} & 90.79 &
  \underline{92.23} & 94.77 &
  90.27 & 92.17 &
  88.53 & 87.73 &
  87.71 & 97.00 &
  94.65 & 95.61 &
  35.16 & 14.79 &
  88.11 & 88.78 &
  88.15 & 75.05 &
  76.57 & 74.94 \\
\textbf{ChuBERT NBDW$_{mixed}$} &
  \underline{94.39} & \textbf{91.93} &
  \textbf{92.73} & 94.22 &
  90.02 & 91.71 &
  86.36 & 85.50 &
  85.73 & 97.77 &
  95.30 & 96.35 &
  34.58 & 12.21 &
  90.36 & 90.94 &
  90.52 & \underline{78.61} &
  79.32 & 78.63 \\ \hline
\textbf{CamemBERT NACHOS$_{small}$} &
  81.44 & 81.39 &
  80.96 & 79.74 &
  78.08 & 78.70 &
  80.59 & 79.88 &
  80.04 & 95.64 &
  91.57 & 92.46 &
  32.87 & 13.76 &
  67.56 & 77.48 &
  71.10 & 55.45 &
  62.34 & 57.43 \\
\textbf{PubMedBERT NACHOS$_{small}$} &
   92.51 & \underline{91.49} & 91.53 &
   \underline{94.95} & \textbf{92.55} & \textbf{93.62} &
   84.73 & 83.80 & 83.85 & 
   97.82 & 96.12 & 96.81 &
   35.88 & 15.21 &
   90.97 & 91.27 & 91.03 &
   \textbf{82.03} & \textbf{81.71} & \textbf{81.73} \\ 
\textbf{CamemBERT NBDW$_{small}$} &
  82.35 & 81.59 &
  81.57 & 78.14 &
  76.38 & 77.12 &
  79.44 & 79.79 &
  79.25 & 95.98 &
  92.11 & 93.18 &
  27.73 & 11.89 &
  53.44 & 73.11 &
  61.75 & 48.71 &
  61.33 & 53.05 \\ \hline
\end{tabular}
\caption{Performance on public biomedical downstream tasks. Best model in bold and second is  underlined.\label{fig:downstreamtaskpublic}}
\end{table*}

\begin{table*}[!b]
\scriptsize
\setlength\extrarowheight{1pt}
\centering
\center
\begin{tabular}{l c c c c c}
\hline
\textbf{} & \textbf{GSD} & \textbf{SEQUOIA} & \textbf{SPOKEN} & \textbf{PARTUT} & \textbf{XNLI} \\ \hline

\textbf{CamemBERT OSCAR 138 GB} & \textbf{98.28} & 98.68 & \underline{97.26} & 97.70 & \textbf{81.94} \\
\textbf{CamemBERT OSCAR 4 GB} & 98.14 & \textbf{99.18} & \textbf{97.57} & \underline{97.86} & \underline{81.76} \\
\textbf{CamemBERT CCNET 4 GB} & \underline{98.18} & \underline{98.92} & 97.20 & \textbf{97.92} & 81.26 \\

\textbf{PubMedBERT} & \multicolumn{1}{r}{96.48} & 96.49 & 90.00 & 93.97 & 73.79 \\
\textbf{ClinicalBERT} & \multicolumn{1}{r}{96.49} & 96.31 & 89.60 & 93.17 & 70.57 \\
\textbf{BioBERT v1.1} & \multicolumn{1}{r}{97.32} & 96.54 & 91.81 & 94.52 & 71.54 \\ \hline

\textbf{DrBERT NACHOS$_{large}$} & 96.94 & 98.05 & 95.92 & 96.54 & 72.18 \\
\textbf{DrBERT NACHOS$_{small}$} & 97.17 & 98.21 & 96.38 & 96.45 & 72.86 \\
\textbf{ChuBERT NBDW$_{small}$} & 96.45 & 97.38 & 94.90 & 95.83 & 69.00 \\
\textbf{ChuBERT NBDW$_{mixed}$} & 97.18 & 98.10 & 96.43 & 96.33 & 72.32 \\ \hline

\textbf{CamemBERT NACHOS$_{small}$} & 97.63 & 96.90 & 91.12 & 94.00 & 71.26 \\
\textbf{PubMedBERT NACHOS$_{small}$} & 97.41 & 98.71 & 95.54 & 97.01 & 77.35 \\ 
\textbf{CamemBERT NBDW$_{small}$} & 97.55 & 96.26 & 89.17 & 91.34 & 72.73 \\ \hline

\end{tabular}
\caption{Performance on public domain-general downstream tasks. Best model in bold and second is
underlined.\label{fig:downstreamtaskgeneral}}
\end{table*}

\subsection{Impact of pre-training strategies}
\label{s:pre-train-impact}


As observed both in Tables~\ref{fig:downstreamtaskprivate} and~\ref{fig:downstreamtaskpublic}, models pre-trained completely from scratch (DrBERT NACHOS and ChuBERT NBDW) tend to produce the best results for both types of data sources and tasks ({\it i.e.} private and public). Indeed, considering the F1-score, they obtain the best results on all private tasks and on almost all public ones (5 tasks out of 7). The two public remaining tasks (MUSCA-DET T2 and QUAERO-MEDLINE) are then better handled using PubMedBERT NACHOS$_{small}$, a model that has already been pre-trained on domain-specific data (biomedical English data) then further pre-trained with our French medical data (NACHOS$_{small}$). 

We also observed that continual pre-training from domain generic models (CamemBERT NACHOS$_{small}$ or CamemBERT NBDW$_{small}$) does not allow reaching the performance of the other specific models, neither of these two models reaching the first or second place (in terms of performance) on any task.

Finally, the baseline models trained on generic data (CamemBERT OSCAR) and those trained on biomedical data in English (PubMedBERT, ClinicalBERT and BioBERT) remain competitive in few biomedical public tasks (CAS POS, FrenchMCQA or MUSCA-DET T2), while none of them are placed in first or second place on private tasks. This seems to highlight the difficulty of private tasks when non-matching data are used.

\subsection{Effect of data}
\label{s:data-effect}


Regarding the amount of data used for pre-training models ($small$ vs. $large$ or $mixed$), results show that, the larger the data are, the better the model performs, no matter the pre-training strategy or the source of data (private or public). However, the difference is very low for most tasks, with ${small}$ systems often being ranked second behind large models, even though they contain half as much data.



We notice a clear dominance of models that were pre-trained on web-based sources, specifically OSCAR and NACHOS, when applied to public tasks.
Indeed, models relying on private NBDW data only achieve the best performance (in terms of F1-score) on the MUSCA-DET T1 task. This trend is not quite observed on private tasks, where NBDW-based models obtain more acceptable or even better performance when mixed with public biomedical data (ChuBERT NBDW$_{mixed}$), as seen in Table~\ref{fig:downstreamtaskprivate}. We believe this discrepancy is mainly due to  the different nature of processed data. 

Finally, we observe that English-based models perform closely to the French-based CamemBERT model. This shows the usefulness of pre-training on domain specific data. For example, better results are obtained with continual pre-training of the PubMedBERT model with our specialized data in French (PubMedBERT NACHOS$_{small}$), corroborating our hypothesis about the effectiveness of cross-language knowledge transfer.

\subsection{Performance on general-domain tasks}
\label{s:general-domain}

Table~\ref{fig:downstreamtaskgeneral} gives the results obtained by all PLMs on general domain downstream tasks. These tasks come from~\citet{Martin_2020} who used them to evaluate the CamemBERT model. The first four are POS tagging tasks (GSD, SEQUOIA, SPOKEN and PARTUT), the last being a natural language inference task (XNLI).

All results of our models decrease in performance on all tasks. The most important drop is for the natural language inference task, with a performance of ChuBERT NBDW$_{small}$ almost 13\% lower than CamemBERT 138 GB. We also observe that the specialized models in English are as efficient as our biomedical models in French. It seems quite clear from the previous observations that specialized models are difficult to generalize to other tasks, but that specialized information captured in one language could transfer to another language.

\section{Conclusion}

In this work, we proposed the first biomedical and clinical Transformer-based language models, based on RoBERTa architecture, for French language. An extensive evaluation study of these specific models has been performed on an aggregated collection of diverse private and public medical tasks. Our open-source DrBERT models improved the state of the art in all medical tasks against both French general model (CamemBERT) and English medical ones (BioBERT, PubMedBERT and ClinicalBERT). In addition, we showed that pre-training on constrained resources (4 GB) of web-crawled medical makes it possible to compete with, and even frequently surpass, models trained with specialized data from medical reports.




Results also highlighted that continual pre-training on an existing domain-specific English model, here PubMedBERT, is a more viable solution than on a French domain-generalist model while targeting French biomedical downstream tasks. It needs to further investigate the performance of this approach using more data, similar to what we have done with DrBERT NACHOS$_{large}$.



The NACHOS corpus as well as the models pre-trained and all the pre-training scripts\footnote{\href{https://drbert.univ-avignon.fr/}{https://drbert.univ-avignon.fr/}} have been publicly released online under a CC0 1.0 open-source license.

\section{Ethical considerations}

Concerning the risks and biases, all the freely available models pre-trained on NACHOS can supposedly be exposed to some of the concerns presented by the work of \citet{10.1145/3442188.3445922} and \citet{https://doi.org/10.48550/arxiv.2105.04054} since some of the NACHOS sub-corpora quality might be lower than expected, specifically for non-governmental sources. When using a BERT-based biomedical language model, potential biases can be encountered including fairness, gendered language, limited representation and temporal correctness.

\section{Limitations}


It is important to mention some limitations of our work. Firstly, it would be wise to evaluate the impact of the tokenizer on the performance of the models to ensure that this is not the main reason for the observed performance gains.

Furthermore, we can not affirm in this study whether the medical domain transfer observed from English to French using continual pre-training on PubMedBERT can be generalized to other languages or other domains.

Finally, it is possible that training a ChuBERT model with more diverse private clinical data and in a larger quantity could have brought notable performance gains on private tasks.

A considerable amount of computational resources was used to conduct this study, since approximately 18,000 hours of GPU computation were used to create the 7 models presented here, as well as about 7,500 hours of GPU for debugging due to technical issues related to model configurations and poor performance, for a total of 25,500 hours. The total environmental cost, according to the Jean Zay supercomputer documentation~\footnote{http://www.idris.fr/media/jean-zay/jean-zay-conso-heure-calcul.pdf} is equivalent to 6,604,500 Wh or 376.45 kg CO2eq based on the carbon intensity of the energy grid mention by BLOOM environmental cost study also made on Jean Zay \cite{luccioni2022estimating}. This makes the present study difficult to reproduce and to transpose to other languages when limited material resources are available.







\section*{Acknowledgments}
This work was performed using HPC resources from GENCI-IDRIS (Grant 2022-AD011013061R1 and 2022-AD011013715) and from CCIPL (Centre de Calcul Intensif des Pays de la Loire). This work was financially supported by ANR AIBy4 (ANR-20-THIA-0011) and Zenidoc. 


\bibliography{anthology,custom}
\bibliographystyle{acl_natbib}


\appendix

\section{Appendix}
\label{sec:appendix}

\subsection{Vocabularies Inter-coverage}

\begin{figure}[H]
\centering
\center
\includegraphics[scale=0.40]{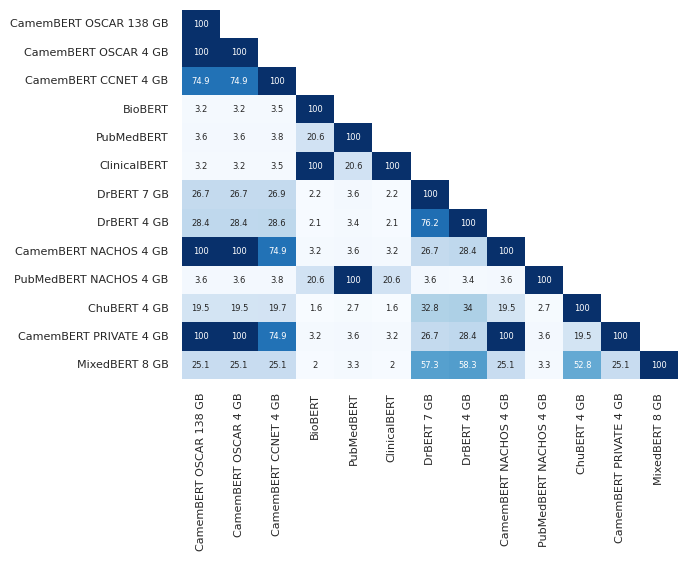}
\caption{Vocabularies Inter-coverage Matrix.}
\label{fig:Inter-Vocab-Coverage}
\end{figure}

As we can see in Figure~\ref{fig:Inter-Vocab-Coverage}, despite having similar performances, some of the models do not share a lot of mutual vocabulary. 


\subsection{Models Stability}

We observe during the evaluation phase that most of the models based on continual pre-training strategy from CamemBERT OSCAR 138 GB are suffering from bad consistency and stability during fine-tuning, which translates into fluctuation in performance between runs.

\begin{figure}[H]
\centering
  \renewcommand{\thesubfigure}{\arabic{row}.\alph{subfigure}}%
  \centering 
  \setcounter{row}{1}%
  \begin{subfigure}[b]{0.24\textwidth}
     \includegraphics[width=\linewidth]{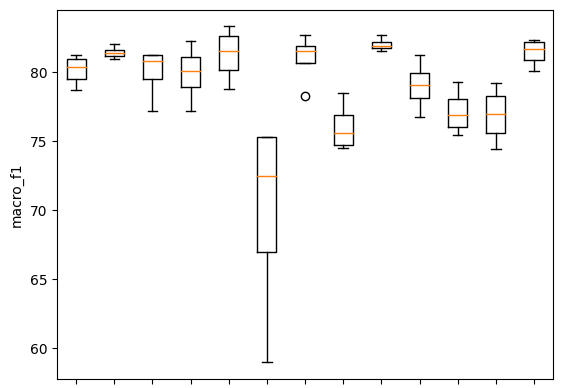}
     \caption{aHF classification}
     \label{fig:c1w}
   \end{subfigure}%
   \begin{subfigure}[b]{0.24\textwidth}
     \includegraphics[width=\linewidth]{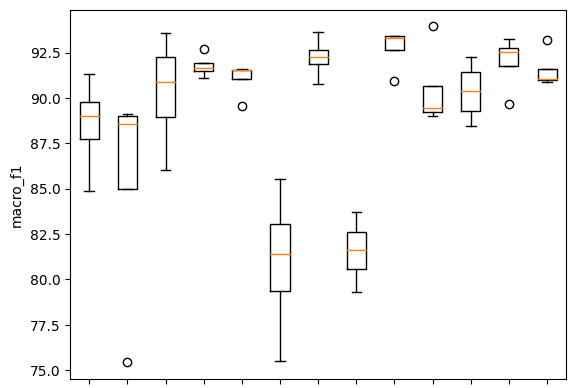}
     \caption{MUSCADET T1}
     \label{fig:c1po}
   \end{subfigure}%

   \stepcounter{row}%
   \begin{subfigure}[b]{0.24\textwidth}
      \includegraphics[width=\linewidth]{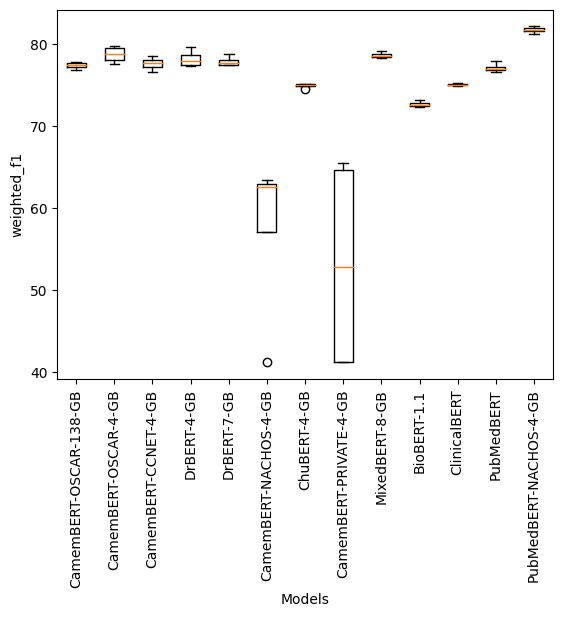}
      \caption{QUAERO MEDLINE}
      \label{fig:c2w}
    \end{subfigure}%
    \begin{subfigure}[b]{0.24\textwidth}
      \includegraphics[width=\linewidth]{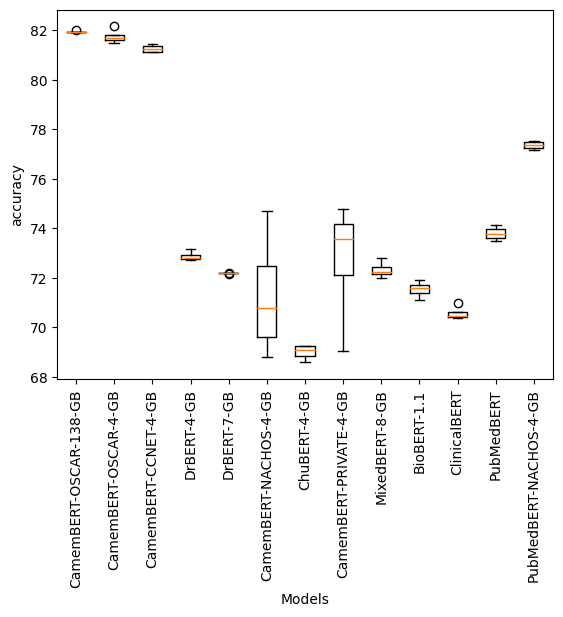}
      \caption{XNLI}
      \label{fig:c2po}
    \end{subfigure}%
     \caption{\label{fig:stabilityscores} Box plot for each model. }
\end{figure}

We also notice during PubMedBERT NACHOS$_{small}$ pre-training that the model loss is globally stable during almost all the duration of the pre-training, until reaching the step 71,000, where the loss fall down until touching down zero at step 72,500.

\begin{figure}[H]
\centering
\center
\includegraphics[scale=0.07]{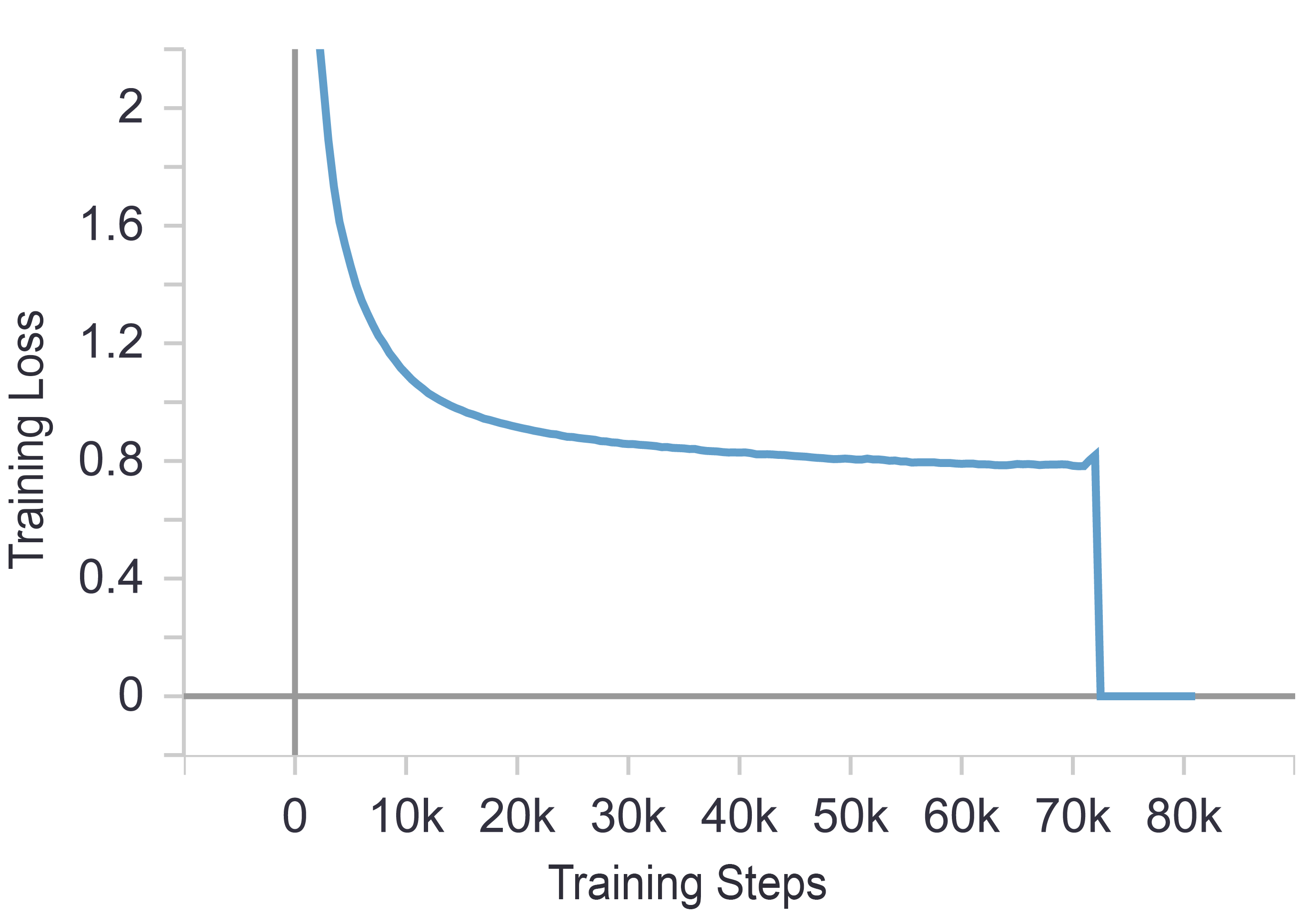}
\caption{PubMedBERT NACHOS$_{small}$ loss.}
\label{fig:Inter-Vocab-Coverage}
\end{figure}




\end{document}